\documentclass[10pt,twocolumn,letterpaper]{article}

\usepackage{wacv}
\usepackage{times}
\usepackage{color}
\usepackage{epsfig}
\usepackage{graphicx}
\usepackage{amsmath}
\usepackage{amssymb}
\usepackage{subfigure}

\usepackage[pagebackref=true,breaklinks=true,letterpaper=true,colorlinks,bookmarks=false]{hyperref}

\wacvfinalcopy 

\def\wacvPaperID{961} 

\ifwacvfinal\fi
\begin{document}

\title{Automatic Head Overcoat Thickness Measure with NASNet-Large-Decoder Net}

\author{Youshan Zhang, Brian D.\ Davison \\
Lehigh University\\
{\tt\small \{yoz217, bdd3\}@lehigh.edu}
\and
Vivien W.\ Talghader, Zhiyu Chen, Zhiyong Xiao, Gary J.\ Kunkel \\
Seagate Technology\\
{\tt\small \{vivien.w.talghader, zhiyu.chen, zhiyong.x.xiao, gary.j.kunkel\}@seagate.com}
}

\maketitle
\ifwacvfinal\thispagestyle{empty}\fi

\begin{abstract}
Transmission electron microscopy (TEM) is one of the primary tools to show microstructural characterization of materials as well as film thickness. However, manual determination of film thickness from TEM images is time-consuming as well as subjective, especially when the films in question are very thin and the need for measurement precision is very high. Such is the case for head overcoat (HOC) thickness measurements in the magnetic hard disk drive industry. It is therefore necessary to develop software to automatically measure HOC thickness. In this paper, for the first time, we propose a HOC layer segmentation method using NASNet-Large as an encoder and then followed by a decoder architecture, which is one of the most commonly used architectures in deep learning for image segmentation. To further improve segmentation results, we are the first to propose a post-processing layer to remove irrelevant portions in the segmentation result. To measure the thickness of the segmented HOC layer, we propose a regressive convolutional neural network (RCNN) model as well as orthogonal thickness calculation methods. Experimental results demonstrate a higher dice score for our model which has lower mean squared error and outperforms current state-of-the-art manual measurement.

\end{abstract}

\section{Introduction}

Transmission electron microscopy (TEM) is a microscopy technique in which a beam of electrons is transmitted through a specimen to form an image. TEM is one of the premier tools to show microstructural characterization of materials \cite{fultz2012transmission}. Differences in material density show up as contrast in a TEM image.

The hard disk drive industry uses TEM to measure head overcoat (HOC) thickness on magnetic recording heads. In a bright-field TEM image, HOC appears as a white layer sandwiched between two darker layers, a magnetic metal layer that is part of the recording head and a capping layer deposited as part of TEM sample preparation. This is because the HOC is comprised predominantly of Diamond-like Carbon (DLC) which has relatively low density whereas the other two layers are of much higher density.

Current state-of-the-art HOC thickness is sub-50A and its accurate and precise measurement is critical for deposition process feedback since HOC thickness has a direct impact on magnetic recording performance. There is currently no automatic method to calculate the thickness of HOC from a TEM image, and manual calculation is time-consuming and highly subjective. Therefore, there is an industrial need to develop an efficient algorithm that can automatically measure HOC thickness.

There are several issues in need of resolution to automatically measure the HOC layer thickness. We do not have segmentation masks already defined for HOC, and there is no standard rule to define the thickness of the HOC layer---the boundary between the HOC layer and the sandwiching layers can be fuzzy, making it challenging to consistently discern, whether by the same operator or different operators. However, even if we have a clear mask of the HOC layer, it is still difficult to determine the HOC thickness due to the curvature and slope of the masked area. Hence, our challenges are: 1)  how to correctly generate a segmentation that separates the HOC from the surrounding material; and, 2) how to accurately measure the thickness of the HOC layer.

Many traditional techniques have been proposed for segmentation, such as threshold methods \cite{tobias2002image,tao2003image,zhang2008image}, region-based methods \cite{freixenet2002yet,ning2010interactive}, genetic methods \cite{hammouche2008multilevel}, level set methods \cite{vese2002multiphase,li2010distance,li2011level} and artificial neural networks \cite{reddick1997automated,awad2007multicomponent,awad2010unsupervised}.

The threshold method is one of the most common and straightforward segmentation methods. It is a region segmentation technique which divides gray values into two or more gray intervals, and chooses one or more appropriate thresholds to judge whether the region meets the threshold requirement according to the difference between the target and the background, and separates the background and the target to produce a binary image. Threshold processing has two forms: global threshold and adaptive threshold. The global threshold only sets one threshold, and the adaptive threshold sets multiple thresholds. The target and background regions are segmented by determining the threshold at the peak and valley of the gray histogram \cite{hu2001automatic,pu2008adaptive}. Level set methods are also widely used in the segmentation task. The basic idea of the level set method for image segmentation is the continuous evolution of curve motion. The boundary of the image is searched until the target contour is found and then the moving curve is stopped. Curves are moved along every three-dimensional section of images to slice different levels of the three-dimensional surface. The level of the obtained closed curves of each layer change over time, and finally get a corresponding shape extraction contour \cite{vese2002multiphase}. In our tasks, the brightness is not consistent from image to image. Therefore, the performance of traditional threshold-based and level-set based methods are not good.

Deep neural networks have also been applied in segmentation and they supersede many traditional image segmentation approaches. Garcia et al.\ presents an overview of deep learning-based segmentation methods \cite{garcia2017review}. There are several models to address the segmentation task. Fully convolutional networks (FCN) \cite{long2015fully} is one of the earliest approaches for deep learning-based image segmentation which performs end-to-end segmentation. FCN is a type of convolutional network for dense prediction that does not require an additional fully connected layer. This method leads to the possibility of segmenting images of any size effectively, and it is much faster than the patch classification method (which divides images into several patches). Almost all of the other more advanced methods follow this architecture. However, there are several limitations of the FCN model, such as the inherent spatial invariance causes the model to fail to take into account  useful global context information and its efficiency in high-resolution scenarios is worse and not available for real-time segmentation. Another difficulty of using CNN networks in segmentation is the existence of pooling layers. The pooling layer not only enlarges the sensing field of the upper convolution layer but also aggregates the background and discards part of the location information. However, the semantic segmentation method needs to adjust the category map accurately, so it needs to retain the location information abandoned in the pooling layer. 

Later the encoder-decoder architecture becomes widely used in segmentation. Segnet \cite{badrinarayanan2017segnet} and U-net \cite{ronneberger2015u} are representative encoder-decoder architectures. This architecture first selects a classification network such as VGG-16, and then removes its full connection layer to produce a low-resolution image representation or feature mapping. This part of the segmentation network is usually called an encoder. A decoder is a complementary part of the network that learns how to decode or map these low-resolution images to the prediction at the pixel level. The difference between different encoder-decoder architectures is the design of the decoder. U-net uses a structure called dilated convolutions and removes the pooling layer structure. Chen et al.~\cite{chen2014semantic} proposed the Deeplab model, which used dilated convolutions and fully connected conditional random field to implement the atrous spatial pyramid pooling (ASPP), which is an atrous version of SPP~\cite{he2015spatial} and can account for different object scales and improve the accuracy.

After getting the segmented HOC layer from a segmentation network, we need to predict the thickness of the HOC layer. Thickness measurement is a regression problem, and deep networks have been developed to solve such tasks. Kang et al.~\cite{kang2014convolutional} proposed a convolutional neural network for estimating image quality, but its architecture is simple, consisting of max pooling and two dense layers. Zhang et al.~\cite{zhang2019regressive} proposed a regressive neural network for electricity prediction, but the input is a vector, not images. Rothe et al.~\cite{rothe2015dex} proposed a regressive VGG-16 network for age prediction. However, a complex neural network may not solve the problem due to  overfitting, since the input to the neural network here is the segmented HOC layer.  Therefore, it is necessary to design a robust regression convolutional neural network model for predicting the thickness of the HOC layer.

\begin{figure*}[h]
\centering
\includegraphics[width=1.8 \columnwidth]{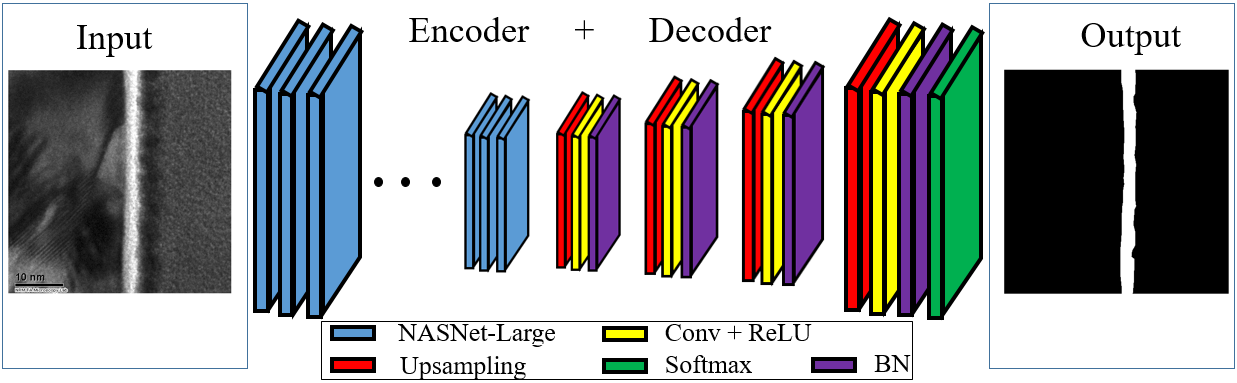}
\caption{The architecture of our NASNet-Large segmentation network. The Encoder consists of the first 414 layers from NASNet-Large model. There are four blocks in the decoder, and each block contains Upsampling, Conv+ReLu, and BN layers. (Convolution (Conv), Batch normalization (BN), Rectified linear units (ReLU).) The final decoder is fed into a softmax layer for HOC layer prediction.  }
\label{fig:Nastnet_segnet}
\end{figure*}

Our contributions are three-fold:
\begin{enumerate}
    \item We are the first to define a NASNet-Large-decoder network to segment TEM head overcoat images;
    \item To remove irrelevant small segments, we are the first to propose a post-processing layer which is able to filter out false segmented sections in prediction images; and, 
    \item We propose an orthogonal calculation method and a regressive convolutional neural network to predict the thickness of the segmented head overcoat layer. Experimental results demonstrate that the proposed model achieves the highest dice and IoU score versus state-of-the-art methods.
\end{enumerate}

This paper is organized as follows: in Sec.~\ref{sec:method}, the NASNet-Large-decoder network, orthogonal distance calculation and regressive networks are summarized; we present the segmentation results and the predicted thickness in Sec.~\ref{sec:results}; in Sec.~\ref{sec:dis}, we discuss the advantages and disadvantage of proposed model and conclude in Sec.~\ref{sec:conclusion}.

\section{Methods}\label{sec:method}
In this section, we first introduce the network to segment the TEM HOC images and then describe the regressive network and concept of an orthogonal distance to calculate the thickness of segmented HOC images.

\subsection{NASNet-Large segmentation network}\label{sec:seg_net}
The NASNet-Large segmentation network contains an encoder and a decoder, which is followed by a classification layer. The architecture is shown in Fig.~\ref{fig:Nastnet_segnet}. There are two significant differences in our model comparing with Segnet, which employs the pre-trained VGG16 network for the encoder. Our Nasnet-Large-decoder net uses the first 414 layers of Nasnet-Large net (which is a network well-trained for ImageNet classification) as the encoder to decompose images \cite{zoph2018learning}. We choose the first 414 layers because the size of the final layer is close to the size of the original image, so it will not lose much information.  If we choose the last layer as the feature extraction layer, it will destroy significant structural information of objects since the last layer will be more suitable for classification, rather than segmentation. We do not use the pre-trained weights but retrain the net using new data to fit Nasnet-Large in our experiment since our dataset is significantly different from ImageNet. 
In addition, the decoder is different and there are no pooling indices in our model since Nasnet-Large net can produce detailed information for the decoder.

An appropriate decoder can upsample its input feature map using the max-pooling layer. The decoding technique is illustrated in Fig.~\ref{fig:Nastnet_segnet}. There are four blocks in the decoder. Each block begins with upsampling which can expand the feature map, followed by convolution and rectified linear units. A batch normalization layer is then applied to each of these maps. The first decoder, which is closest to the last encoder can produce a multi-channel feature map. This is similar to Segnet, which can generate a different number of sizes and channels as their encoder inputs. 
The final output of the last decoder layer is fed to a trainable soft-max classifier which produces a $K$ channel image of probabilities where $K$ is the number of classes (two in our problem). The predicted segmentation corresponds to the class with maximum probability at each pixel.

\subsection{Orthogonal calculation of HOC thickness}\label{sec:orth}
\begin{figure}[t]
\centering
\includegraphics[width=0.8 \columnwidth]{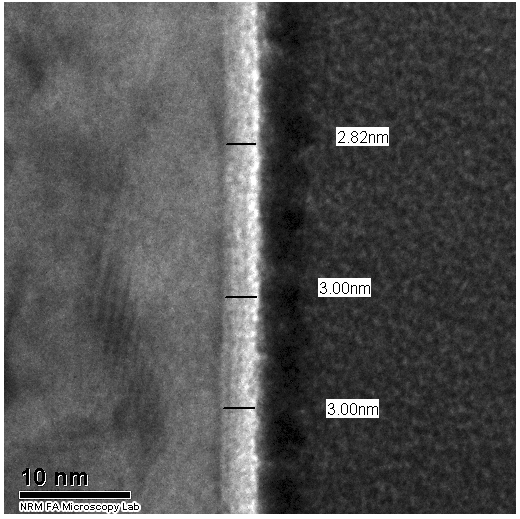}
\caption{The existing three-line thickness measurement of the HOC layer. However, the three-line approach is manually labeled, and consists of horizontal lines, which will overestimate the thickness if within a slope or curve.}
\label{fig:three_lines}
\end{figure}

\begin{figure}[t]
\centering
\includegraphics[width=0.7 \columnwidth]{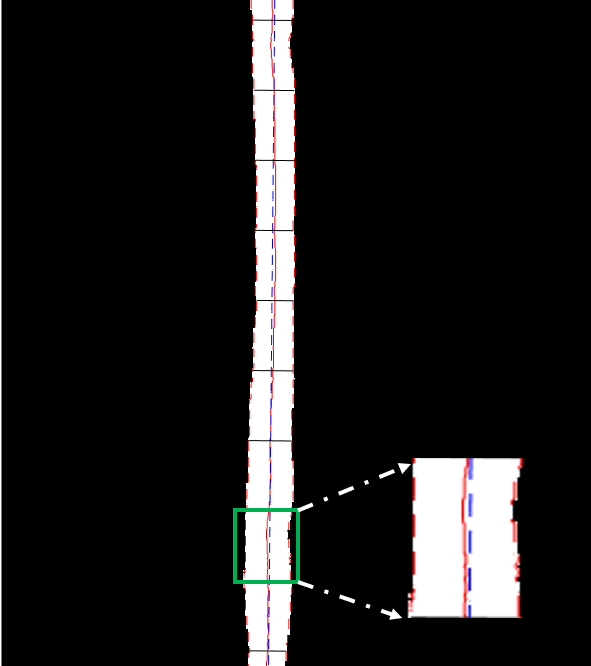}
\caption{The orthogonal distance of the HOC layer. Two red dashed lines are the boundaries of the HOC layer; the middle red line consists of midpoints of boundary points, and the blue dashed line is the regression line. Several black lines are the orthogonal distance of the HOC layer. Lower right shows magnified green box.}
\label{fig:orth_dis}
\end{figure}

The current measurement technique for HOC layer thickness is a manually labeled image with three lines as shown in Fig.~\ref{fig:three_lines}. From the lengths of the three lines, we calculate the mean thickness and standard deviation of the HOC layer. However, there are two obvious shortcomings of this method. One is that it only measures three lines, not all lines of the HOC layer. Secondly, the three lines are horizontal, which may not measure the orthogonal distance of HOC layer. If there is a slope in the HOC layer, the horizontal three-line mean will produce a larger distance than the real thickness. Therefore, to calculate thickness of the segmented HOC layer from Sec.~\ref{sec:seg_net}, and capture the physical meaning of the measurement, we also propose to calculate an orthogonal distance. We need to give a formal definition of thickness of the HOC layer. As shown in Fig.~\ref{fig:orth_dis}, ten black lines sample the thickness of the HOC layer, and are perpendicular to the middle regression line. There are more than one hundred orthogonal lines in the segmented HOC mask (i.e., one at each pixel); we only show ten of them. The thickness of HOC is defined as the mean of all lines which are perpendicular to middle regression line. The standard deviation (SD) represents the variation in thickness across all orthogonal lines.

\begin{figure*}[h]
\centering
\includegraphics[width=2 \columnwidth]{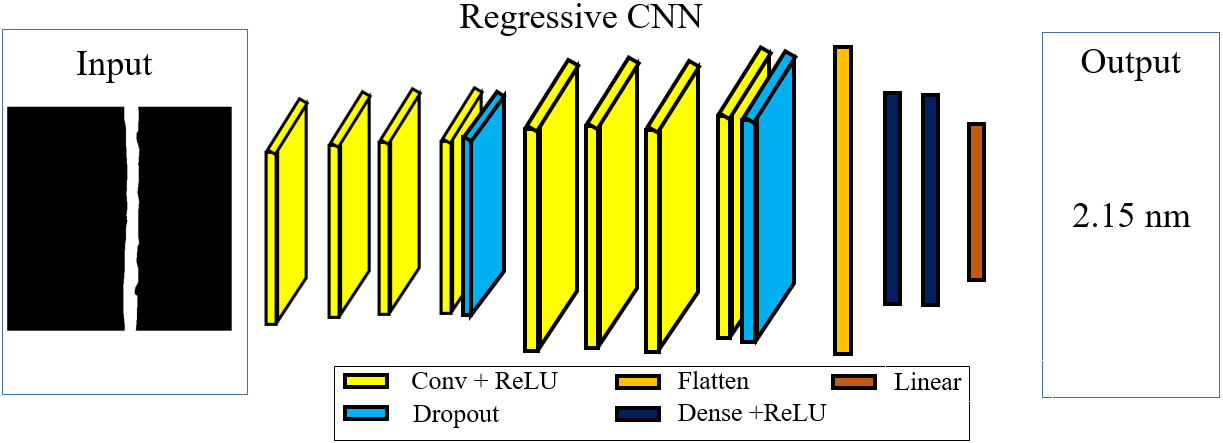}
\caption{The architecture of the thickness regression network. The input is from the NASNet-Large segmentation network. The thickness regression network primarily consists of two blocks, and each block includes Conv + ReLU and Dropout layers. It is then followed by a Flatten layer. Finally, it ends with a linear activation layer. The output is the thickness of the HOC layer.}
\label{fig:Regression-net}
\end{figure*}

\subsection{Regressive thickness prediction net}
However, orthogonal calculation of thickness heavily depends on the boundary of segmented mask. To alleviate this issue, we propose a regressive CNN (RCNN) model, which is described in Fig.~\ref{fig:Regression-net}. The input is the segmented mask from Sec.~\ref{sec:seg_net}, which contains all information about the HOC layer within the images instead of simple boundary of the image, and the output is mean thickness of HOC layer. There are two major blocks in the RCNN model. Each of them consists of three convolutional layers and is followed by one dropout layer. After that a flatten layer expands all features in its front dropout layer into a vector. Two dense and activation layers follow. The regression network ends with a linear activation to predict the thickness of the HOC layer. During the training stage of the RCNN model, the labeled thickness is from the proposed orthogonal calculation method.

There are two advantages of our RCNN model. First, the RCNN model is not complex and easy to train. If the architecture is too complex, it may lead to an overfitting problem. One reason is the input is the segmented HOC layer, which does not contain too much information. Hence, a complex neural network is unnecessary. Another advantage of the RCNN model is that it can give a prediction with less error. In the training stage, the RCNN model is fed with the segmented mask and outputs orthogonal calculation distance. The loss function is mean squared error (MSE), which is defined in Eq.~\ref{eq:mse}. 
\begin{equation}\label{eq:mse}
    \text{MSE}=\frac{1}{n} \sum_{i=1}^n (Y_i-\hat{Y_i})^2,
\end{equation}
where $Y_i$ is the prediction value, and $\hat{Y_i}$ is the orthogonal distance measurement as shown in Sec.~\ref{sec:orth}. Therefore, we minimize the error between orthogonal distance using the predicted HOC mask and the orthogonal thickness. This further drives the prediction mean thickness to be close to the orthogonal thickness. 

Unfortunately, a deep network is often difficult to trust given its opacity.  Domain experts prefer a clear measurement.  However, the proposed RCNN model can be validated, if needed, by the orthogonal calculation.

\section{Results}\label{sec:results}
\subsection{Datasets}
\begin{figure}[t]
\centering
\includegraphics[width=1 \columnwidth]{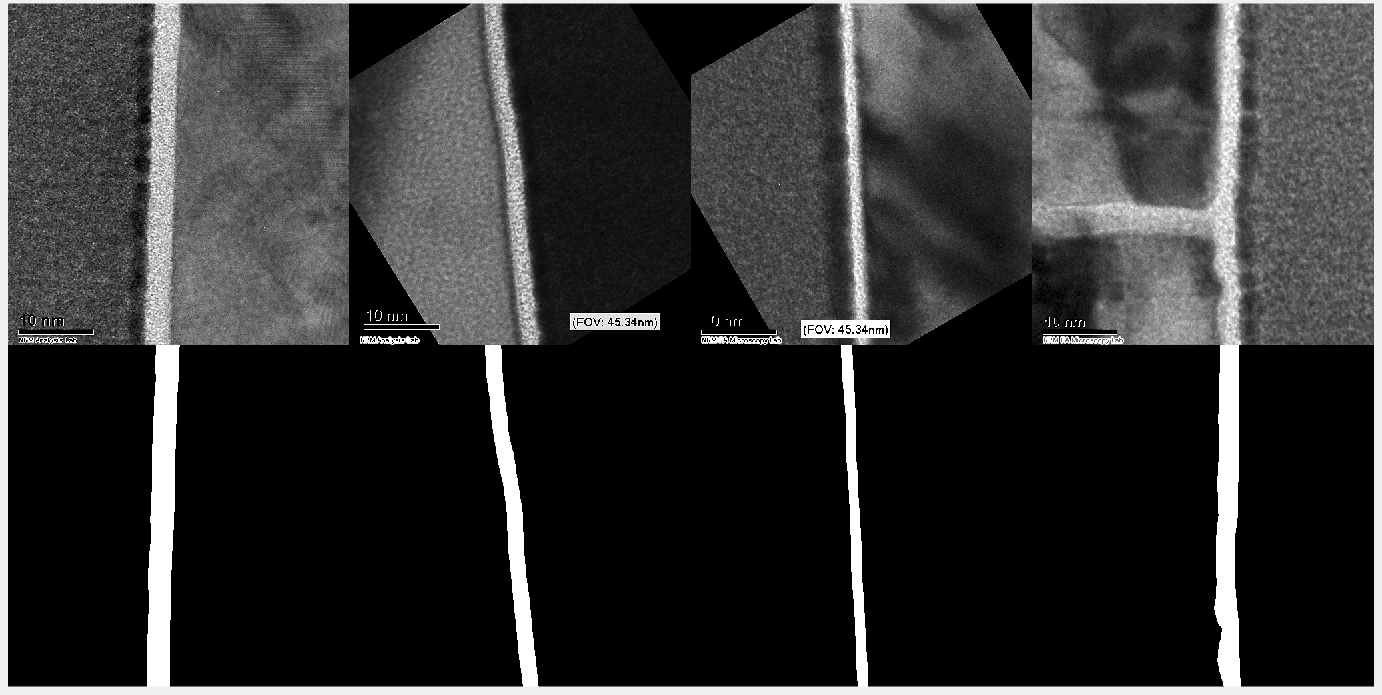}
\caption{Four TEM HOC images and their corresponding HOC segmentations.}
\label{fig:exam}
\end{figure}

Our modeling was trained with data, that are collected from three different dates, differing in the composition of the HOC material.
Due to a limited sample size of 364 images, we use five-fold cross-validation  to estimate the performance of our model. We normalize pixel values of the image into a scale between 0 and 1. Since we do not have ground truth for the mask of HOC layer, the TEM image was first marked by an expert using  thickness measurement software (an internally developed package based on Zhang et al.~\cite{zhang2018mask}) to manually segment the mask and output rough thickness.  Fig.~\ref{fig:exam} shows four TEM images with their corresponding segmentation labels. The middle two examples in Fig.~\ref{fig:exam} cause difficulty with the HOC layer segmentation; there is a FOV (field of view) sticker in the third image, and there is an extra horizontal layer in the fourth image. Due to these reasons, the traditional threshold-based methods failed in segmenting these three images.

Our experiments are implemented within Keras, and our network was trained on an NVIDIA TITAN XP equipped with 12 GB of memory in order to exploit its computational speed. The network parameters were set to: Batch size: 4, Step size: 5, Number of epochs: 1000.

\subsection{Segmentation results visualization}
\begin{figure}[h]
\centering
\includegraphics[width=1 \columnwidth]{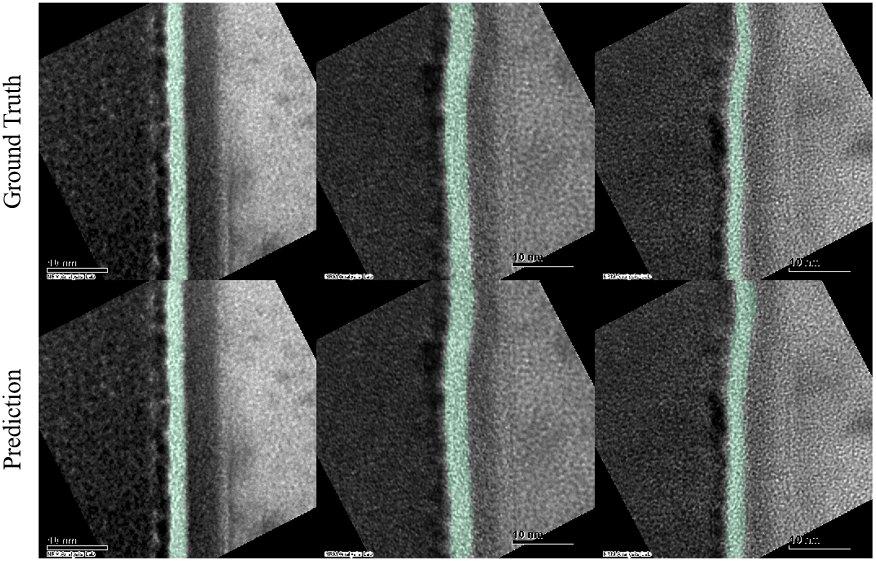}
\caption{Ground truth vs.\ prediction results. The first row is the ground truth mask overlay with the raw image, and the second row is the prediction result overlaid with the raw image.}
\label{fig:good}
\end{figure}

Fig.~\ref{fig:good} compares the predicted segmented HOC image with the ground truth image in the test dataset. The prediction image is close to the real mask, which qualitatively demonstrates the high performance of our model.


\subsection{Post-processing layer }
\begin{figure}[t]
\centering
\includegraphics[width=1 \columnwidth]{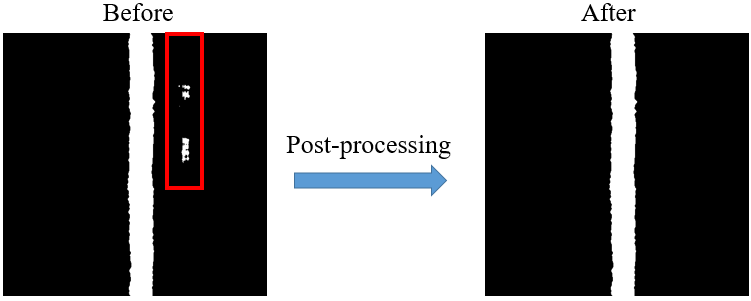}
\caption{The post-processing of prediction result. On the left is the prediction result from the proposed network, and on the right is the result after using our post-processing layer. The red box outlines an irrelevant feature.}
\label{fig:post}
\end{figure}

However, there are some areas in the image which are not the true HOC layer in the prediction result. We then propose a post-processing layer, which can filter the irrelevant parts of the prediction. The first step of post-processing layer is to classify the predicted image into several parts\footnote{The major step of post-processing step is using the {\em connectedComponentsWithStats} function in OpenCV.}, and we then simply select the largest areas (the HOC area) as the final segmented HOC layer. As shown in the left image in Fig.~\ref{fig:post}, the red box highlights erroneous predictions of the image (false negatives). After post-processing, the red box is removed. Therefore, the prediction result can be improved if we filter out such irrelevant parts.

\begin{figure}[t]
\centering
\includegraphics[width=1 \columnwidth]{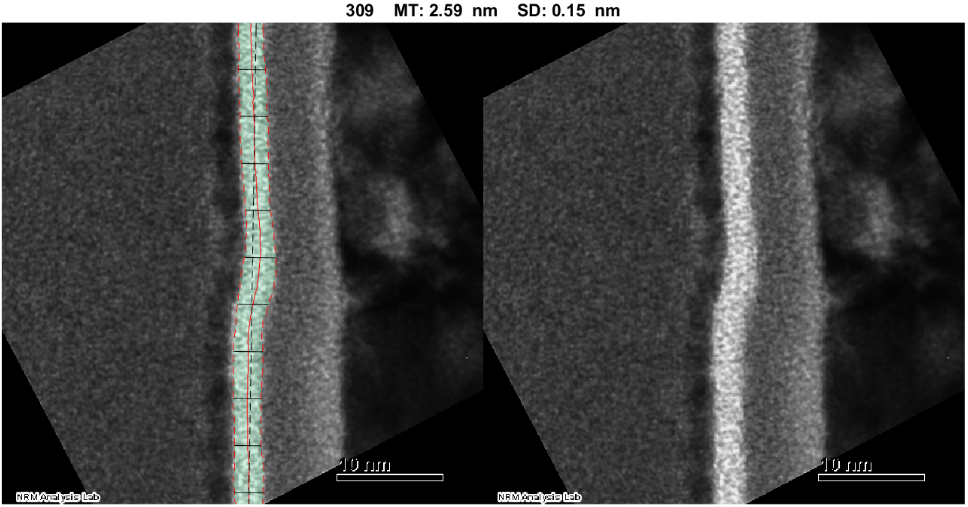}
\caption{Sample output image. Left: the predicted result of our model; right: the original image. The segmented mask is perfectly overlapped with the original image (MT: mean thickness, SD: standard deviation).}
\label{fig:Output}
\end{figure}

\subsection{Metrics}
To evaluate the performance of our NASNet-Large segmentation network, we use the widely-used Dice coefficient index to indicate the goodness of the segmentation results. Furthermore, we also report the IoU score. The two metrics are defined in the following formulas:
\begin{equation*}
    Dice=2 \times \frac{|A \cap  B|}{|A|+|B|}, \ \ \ IoU=\frac{|A \cap B|}{|A \cup B|},
\end{equation*}
where $A$ is ground truth mask, and $B$ is the prediction mask. We also compare our results with performance of state-of-the-art methods.


\begin{table}[h]
\small
\begin{center} 
\caption{Five-fold cross-validation segmentation results}
 \setlength{\tabcolsep}{+3.3mm}{
\begin{tabular}{|c|cc|c|c|c|c|c|c|c|c|c|}
\hline \label{tab:R2}
\bf Method &  \bf IoU & \bf Dice score \\
\hline
U-Net  & 0.80  & 0.85 \\
Deep-Lab  & 0.83  & 0.89 \\
Segnet  & 0.78  & 0.83 \\
\hline
\hline
{\bf NastnetLarge-net}   & 0.86  & 0.92 \\
{\bf NastnetLarge-net-post}   & \textbf{0.89}  & \textbf{0.94} \\
\hline
\end{tabular}}
\end{center}
\end{table} 

\begin{figure}[h]
\centering 
\subfigure[Training with 0.0929 MSE]{\label{fig:va}
\includegraphics[width=0.9\columnwidth]{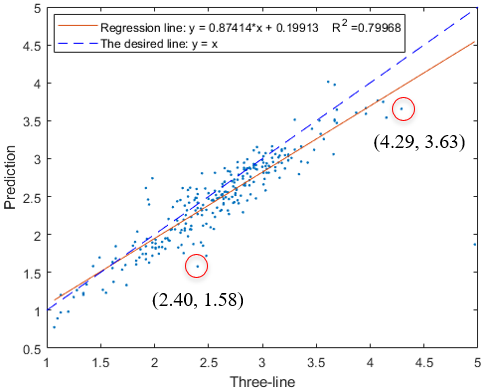}
}
\subfigure[Test with 0.0719 MSE]{\label{fig:te}
\includegraphics[width=0.92\columnwidth]{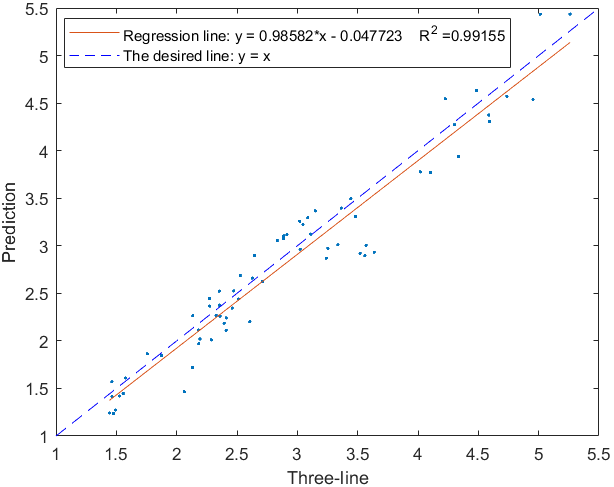}
}
\caption{Comparison of the existing three-line method and our predication in training and test datasets.  }
\label{fig:sta_com}
\end{figure}

Tab.~\ref{tab:R2} lists the comparison results of different models (the metrics are only reported on HOC layer and not the background area). Our NASNet-Large-Decoder has a higher IoU and Dice score than other models. We also observe that the method including post-processing outputs the highest scores, which illustrates that the post-processing layer is useful in our segmentation task.

We use MSE to evaluate the performance of RCNN model and traditional orthogonal distance calculation as defined in Eq.~\ref{eq:mse}. As shown in Tab.~\ref{tab:thick}, the MSE of RCNN model is less than the other three methods, which implies that RCNN model gives a closer prediction to orthogonal measurement than Segnet RCNN and NASNet-Large RCNN. Differing from original Segnet and NASNet-Large model, we replace the final softmax layer with a regressive layer to form the Segnet RCNN and NASNet-Large RCNN model. Hence, these two models are able to predict the thickness. Fig.~\ref{fig:Output} shows the final output of our thickness measurement tool. The left image is the segmented HOC layer overlay with original image, and it also shows orthogonal distance. We also add the file name, mean thickness and standard deviation of the thickness on the top of the image.

\begin{table}[h]
\small
\begin{center} 
\caption{Thickness prediction results comparison}
 \setlength{\tabcolsep}{+7.3mm}{
\begin{tabular}{|c|c|c|c|c|c|c|c|c|c|c|c|}
\hline \label{tab:thick}
Methods &  MSE \\
\hline
\textbf{Regressive CNN}  & \textbf{0.0089} \\
\hline
Three-line measurement  & 0.1524   \\
\hline
Segnet RCNN  & 0.1792   \\
\hline
NASNet-Large RCNN  & 0.2023  \\
\hline
\end{tabular}}
\end{center}
\vspace{-.2in}
\end{table} 

\section{Discussion}\label{sec:dis}
In this section, we first analyze predicted HOC thickness and then discuss the merits and demerits of our model.

\subsection{Statistical analysis}

\begin{figure}[h]
\centering
\subfigure[(2.40, 1.58)]{\label{fig:p1}
\includegraphics[width=0.42\columnwidth]{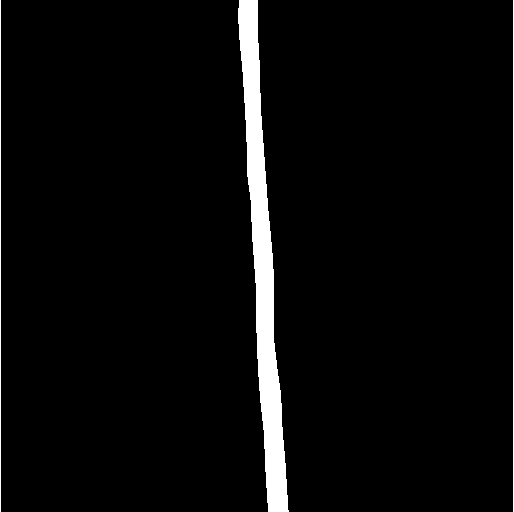}}
\centering 
\subfigure[(4.29, 3.63)]{\label{fig:p2}
\includegraphics[width=0.42\columnwidth]{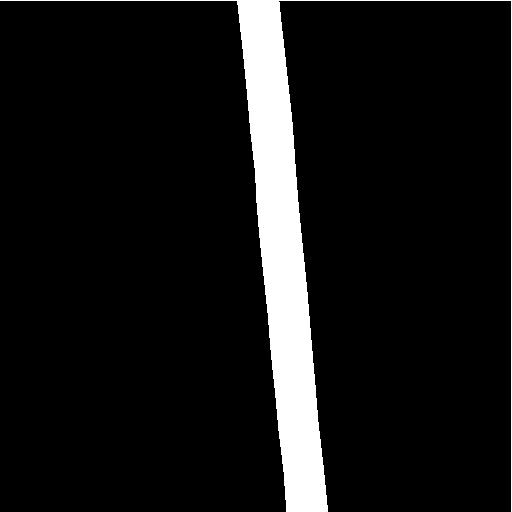}
}
\caption{Two outlier images, which correspond to the highlighted points in Fig.~\ref{fig:va}. Labels are (X,Y) pairs.  X-value is the thickness from three-line measurement and y-value is the prediction value from orthogonal distance. }
\label{fig:two_points_mask}
\end{figure}

\begin{figure}[t]
\centering
\includegraphics[width=0.9 \columnwidth]{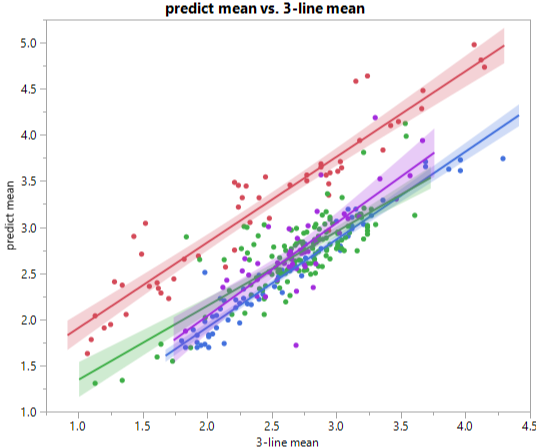}
\caption{Linear regression fit of predicted thickness and the three -line thickness (Different color represents data are collected from different dates).}
\label{fig:fit}
\end{figure}

\begin{figure}[h]
\centering 
\subfigure[Deep learning prediction]{\label{fig:deep_XRF}
\includegraphics[width=0.9\columnwidth]{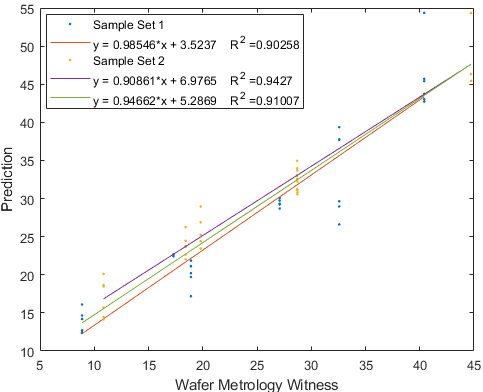}}
\centering 
\subfigure[Human measurement ]{\label{fig:three-XRF}
\includegraphics[width=0.9\columnwidth]{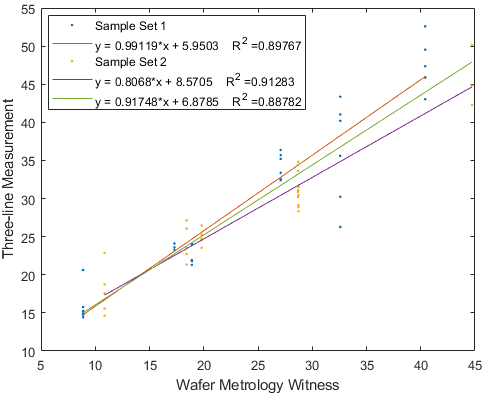}
}
\caption{TEM human 3-line measurement and TEM deep learning prediction compared against measurements made with witness wafer metrology. Deep learning prediction gives more accurate results than human three-line measurement.  
}
\label{fig:XRF}
\end{figure}

As shown in Fig.~\ref{fig:sta_com}, we compare the three-line measurements with orthogonal distance from predicted HOC mask using linear regression in training and test datasets, respectively. The blue dashed lines in the figure indicate the desired line if prediction results are the same as the human three-line measurements. The regression slope from training and test are 0.79968 and 0.99155, respectively, which demonstrate a strong linear correlation between the human measurement and deep learning method. Especially noticable is the test dataset, in which the prediction is almost the same as human measurement. We also calculate the mean squared error (MSE) in Eq.~\ref{eq:mse} to indicate the error between deep learning prediction and human measurement. Although there are some outliers in Fig.~\ref{fig:va}, eg., (2.40, 1.58) and (4.29, 3.63) (the x-value is three-line result and y-value is prediction results from our model), the overall performance of the orthogonal measurement method is closer to human measurement. The three-line method has a higher value than deep learning prediction in these two off-line points. We further check the prediction mask images of these two points as shown in Fig.~\ref{fig:two_points_mask}. We observe that there is a slope in these two prediction HOC layer. Remember, the three-line is measured using horizontal lines, which will lead to a higher value than the real distance if there is a slope in the boundary of mask images. However, we calculate orthogonal distance, which represents the real thickness of the HOC layer. Therefore, we can conclude that our method can be applied in thickness measurements, and it will lead to better results than human three-line measurement. 

Fig.~\ref{fig:fit} describes overall performance in the training and test in terms of three different data collection dates using the mean thickness and standard deviation. We further compare the prediction thickness from our model and the three-line approach with measurements made on witness wafers included in the same HOC depositions as the product heads that were measured with TEM. The witness wafers were measured using metrology calibrated using an amalgamation of techniques, such as AFM (Atomic Force Microscopy), ellipsometry, and XRR (X-Ray Reflectometry).  

From Fig.~\ref{fig:XRF}, we can find that the deep learning-based model has slightly worse results in sample set \#1 since the slope is smaller than the three-line measurement (0.98546 vs.\ 0.99119 ). However, the performance in sample set \#2 is significantly higher than the three-line measurement (0.90861 vs.\ 0.8068). 
Also, the overall prediction of our model is better than human measurement (as shown in the green lines in Fig.~\ref{fig:XRF} ). Therefore, we can conclude that our model has a better performance than traditional manual measurement.

\begin{figure}[t]
\centering
\includegraphics[width=1 \columnwidth]{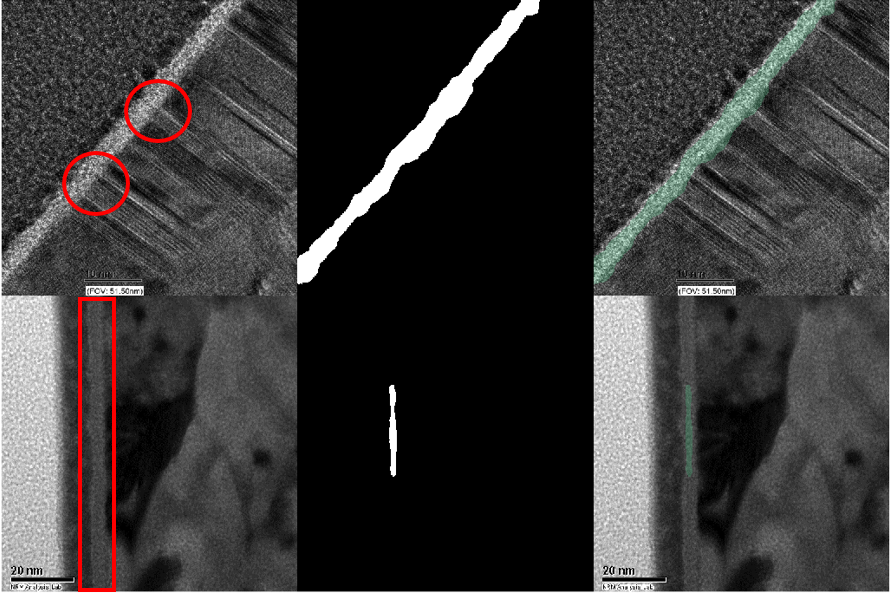}
\caption{Two failures in NASNet-Large-Decoder model. The first row of image is caused by the fuzzy boundary and the second row of image is caused by the unusually dark HOC layer.}
\label{fig:bad}
\end{figure}


Although the RCNN model gives more accurate results than three-line measurement (as shown in Tab.~\ref{tab:thick}), we did not use it to predict thickness of test datasets. 
The crucial reason is that customers require a comparison image for thickness measure as shown in Fig.~\ref{fig:Output}. But RCNN model cannot output images with actual calculation of distance (i.e., cannot output an image as in the left of Fig.~\ref{fig:Output}).  Hence it is not used on test datasets. However, RCNN model is a back-up model if the orthogonal distance failed.  

One of the distinct advantages of our model is that it achieves a higher Dice and IoU score. And there are two reasons: the designed NASNet-Large segmentation net is suitable for TEM image segmentation, and the post-processing layer filters out the unnecessary parts in the image, which improves the segmented results.

Although our model achieves a 0.94 Dice score, it still fails in some cases. As shown in Fig.~\ref{fig:bad}, we observe that segmentation results are worse in these two situations. There are two reasons; one is that there is no similar image in the training dataset, which leads to worse performance. Second is that our model is not robust enough to deal with some unusual images. Comparing with normal training images as shown in Fig.~\ref{fig:exam}, two worse raw images (in Fig.~\ref{fig:bad}) are either too light (first row) or too dark (second row), and it is even difficult for humans to distinguish the HOC layer and background. Also, our model has limited sample size (less than 400 images), and so while in the validation and test datasets the MSE is small, our model could not cover all the cases, e.g., in Fig.~\ref{fig:bad}. Therefore, including more samples in training would help to get a more robust model.
In addition, a limitation of the NASNet-Large-Decoder network is that it needs a large memory to train the model. Therefore, designing a more efficient architecture is a direction for future work.

\section{Conclusion}\label{sec:conclusion}

In this paper, we are the first to present a TEM HOC layer segmentation using the NASNet-Large-decoder architecture, and we get an accurate segmentation with 0.94 Dice score. A post-processing layer is employed to remove the unnecessary part in the prediction map. We then propose an  orthogonal thickness calculation method as well as a regressive convolutional neural network (RCNN) model to measure the thickness of the segmented HOC layer.  These advances enable manufacturers to replace a manual method with a much more accurate automated process. 

Our model can be further improved with a more robust encoder and decoder model, 
to enable it to be applied in a wider variety of cases.
The RCNN model can also be enhanced by predicting both the mean thickness and the standard deviation of the mean thickness.

{\small
\bibliographystyle{ieee}
\bibliography{egbib}
}

\end{document}